%% file: main.tex
\documentclass[runningheads]{llncs}
\usepackage{graphicx}
\usepackage{comment}
\usepackage{amsmath,amssymb} 
\usepackage{color}

\usepackage{times}
\usepackage{epsfig}
\usepackage[utf8]{inputenc} 
\usepackage[T1]{fontenc}    
\usepackage{hyperref}       
\hypersetup{colorlinks=true,allcolors=black}
\usepackage{url}            
\usepackage{booktabs}       
\usepackage{amsfonts}       
\usepackage{nicefrac}       
\usepackage{microtype}      
\usepackage[dvipsnames]{xcolor}
\usepackage{wrapfig}
\usepackage{transparent}
\usepackage{multirow}

\graphicspath{{./figures/}}

\newcommand{\name} {RF-Diary}
\newenvironment{Itemize}%
{\begin{itemize}%
\setlength{\itemsep}{0pt}%
\setlength{\topsep}{0pt}%
\setlength{\partopsep}{0pt}%
\setlength{\parskip}{0pt}}%
{\end{itemize}}
\begin{document}
\pagestyle{headings}
\mainmatter
\def\ECCVSubNumber{2978}  

\title{In-Home Daily-Life Captioning Using Radio Signals}

\titlerunning{In-Home Daily-Life Captioning Using Radio Signals}
%
\author{Lijie Fan\thanks{Indicates equal contribution. Correspondence to Tianhong Li <tianhong@mit.edu>.} \and Tianhong Li$^{\ast}$ \and Yuan Yuan \and Dina Katabi}
%
\authorrunning{Lijie Fan, Tianhong Li, Yuan Yuan and Dina Katabi}
%
\institute{MIT CSAIL}
\maketitle

\begin{abstract}

\input{source/abstract}

\end{abstract}
\input{source/intro-5}
\input{source/related-2}
\input{source/rf}
\input{source/method-2}
\input{source/experiment-3}

\input{source/conclusion}

\bibliographystyle{splncs04}
\bibliography{egbib}

\end{document}

%% file: source/abstract.tex
This paper aims to caption daily life --i.e., to create a textual description of people's activities and interactions with objects in their homes.  Addressing this problem requires novel methods beyond traditional video captioning, as most people would have privacy concerns about deploying cameras throughout their homes. We introduce \name, a new model for captioning daily life by analyzing the privacy-preserving radio signal in the home with the home's floormap. \name\ can further observe and caption people's life through walls and occlusions and in dark settings. In designing \name, we exploit the ability of radio signals to capture people's 3D dynamics, and use the floormap to help the model learn people's interactions with objects. We also use a multi-modal feature alignment training scheme that leverages existing video-based captioning datasets to improve the performance of our radio-based captioning model. Extensive experimental results demonstrate that \name\ generates accurate captions under visible conditions. It also sustains its good performance in dark or occluded settings, where video-based captioning approaches fail to generate meaningful captions.\footnote{For more information, please visit our project webpage: \href{http://rf-diary.csail.mit.edu}{\color{magenta}http://rf-diary.csail.mit.edu}}

%% file: source/intro-5.tex
\section{Introduction}
Captioning is an important task in computer vision and natural language processing; it typically generates language descriptions of visual inputs such as images or videos~\cite{venugopalan2015sequence,yao2015describing,fakoor2016memory,baraldi2017hierarchical,yu2016videornn,pan2016hierarchical,song2017hierarchical,wu2018interpretable,hu2019hierarchical,ranzato2015sequence,pasunuru2017reinforced,wang2018video}. This paper focuses on {\it in-home daily-life captioning}, that is, creating a system that observes people at home, and automatically generates a transcript of their everyday life. Such a system would help older people to age-in-place. Older people may have memory problems and some of them suffer from Alzheimer's. They may forget whether they took their medications, brushed their teeth, slept enough, woke up at night, ate their meals, etc. Daily life captioning enables a family caregiver, e.g., a daughter or son, to receive text updates about their parent's daily life, allowing them to care for mom or dad even if they live away, and providing them peace of mind about the wellness and safety of their elderly parents. More generally, daily-life captioning can help people track and analyze their habits and routine at home, which can empower them to change bad habits and improve their life-style. 

\begin{figure*}[t]
\begin{center}
\includegraphics[width=1.0\linewidth]{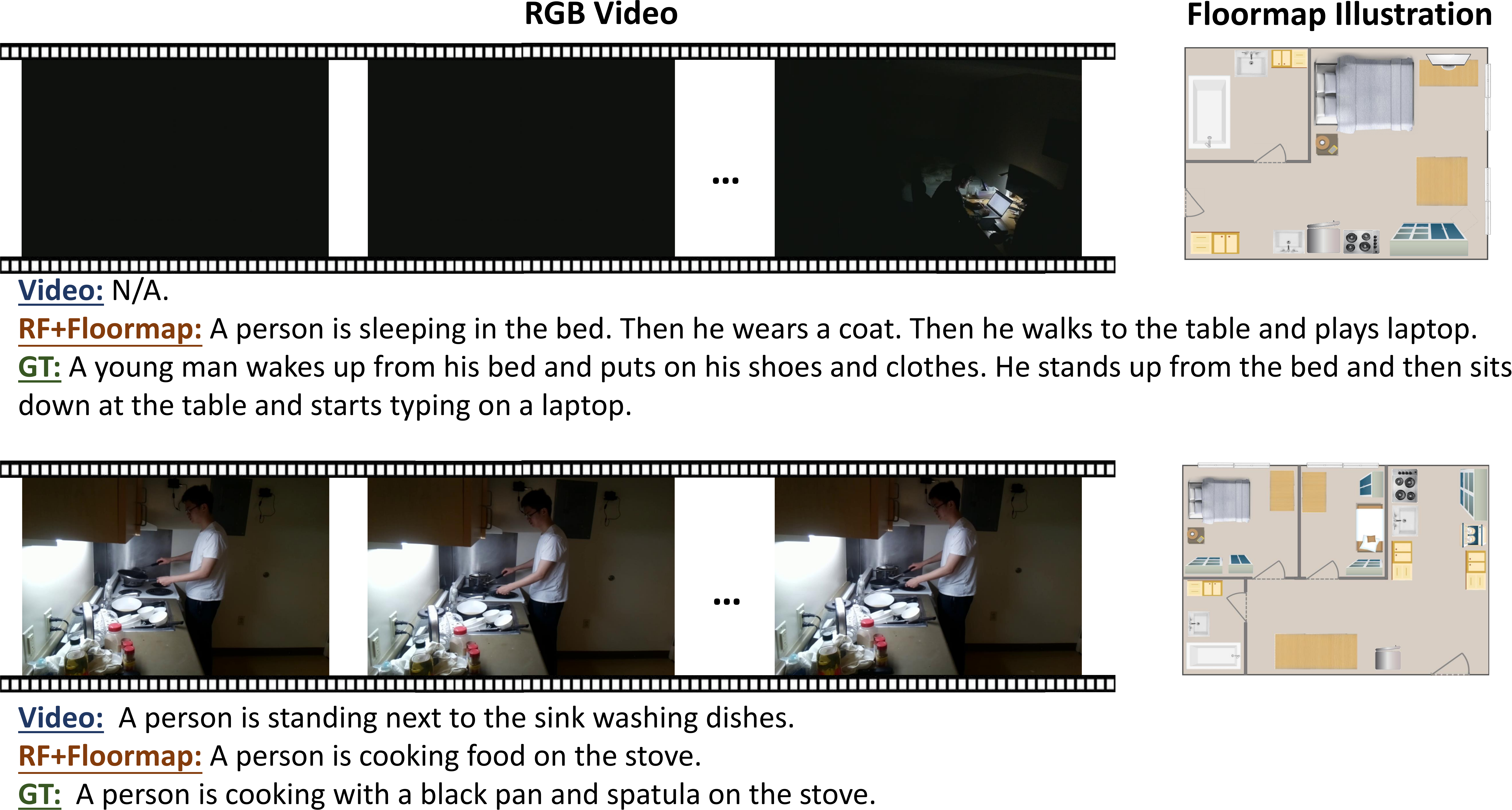}
\end{center}
\vskip -15 pt
\caption{\footnotesize{Event descriptions generated from videos and RF+Floormap. The description generated from video shows its vulnerability to poor lighting and confusing images, while \name\ is robust to both. The visualization of floormap shown here is for illustration.  The representation used by our model is person-centric and is described in detail in section~\ref{sec:floormap}.
}}\label{fig:teaser}
\vskip -15 pt
\end{figure*}

But how do we caption people's daily life? One option would be to deploy cameras at home, and run existing video-captioning models on the recorded videos. However, most people would have privacy concerns about deploying cameras at home, particularly in the bedroom and bathroom. Also, a single camera usually has a limited field of view; thus, users would need to deploy multiple cameras covering different rooms, which would introduce a significant overhead. Moreover, cameras do not work well in dark settings and occlusions, which are common scenarios at home. 

To address these limitations, we propose to use radio frequency (RF) signals for daily-life captioning. RF signals are more privacy-preserving than cameras since they are difficult to interpret by humans. Signals from a single RF device can traverse walls and occlusions and cover most of the home. Also, RF signals work in both bright and dark settings without performance degradation. Furthermore, the literature has shown that one can analyze the radio signals that bounce off people's bodies to capture people's movements~\cite{adib20143d,adib2013see}, and track their 3D skeletons~\cite{zhao2018rf}.


However, using RF signals also introduces new challenges, as described below: 

\begin{Itemize}
\item {\bf Missing objects information:} RF signals do not have enough information to differentiate objects, since many objects are partially or fully transparent to radio signals. Their wavelength is on the order of a few centimeters, whereas the wavelength of visible light is hundreds nanometer~\cite{barbrow1964international}. Thus, it is also hard to capture the exact shape of objects using RF signals. 
\item {\bf Limited training data:} Currently, there is no training dataset that contains RF signals from people's homes with the corresponding captions. Training a captioning system typically requires tens of thousands of labeled examples. However, collecting a new large captioning dataset with RF in people's homes would be a daunting task.
\end{Itemize}

In this paper, we develop \name, an RF-based in-home daily-life captioning model that addresses both challenges. To capture objects information, besides RF signals, \name\ also takes as input the home floormap marked with the size and location of static objects like bed, sofa, TV, fridge, etc. Floormaps provide information about the surrounding environment, thus enabling the model to infer human interactions with objects. Moreover, floormaps are easy to measure with a cheap laser-meter in less than 10 minutes (Section~\ref{sec:floormap}). Once measured, the floormap remains unchanged for potentially years, and can be used for all future daily-life captioning from that home.


\name\ proposes an effective representation to integrate the information in the floormap with that in RF signals. It encodes the floormap from the perspective of the person in the scene. It first extracts the 3D human skeletons from RF signals as in~\cite{zhao2018rf} and then at each time step, it shifts the reference system of floormap to the location of the extracted skeleton, and encodes the location and orientation of each object with respect to the person in the scene. This representation allows the model, at each time step, to focus on various objects depending on their proximity to the person.

To deal with the limited availability of training data, we propose a multi-modal feature alignment training scheme to leverage existing video-captioning datasets for training \name. To transfer visual knowledge of event captioning to our model, we align the features generated from \name\ to the same space of features extracted from a video-captioning model trained on existing large video-captioning datasets. Once the features are aligned, we use a language model to generate text descriptions.

\autoref{fig:teaser} shows the performance of our model in two scenarios. In the first scenario, a person wakes up from bed, puts on his shoes and clothes, and goes to his desk to work on his laptop. \name\ generates a correct description of the events, while video-captioning fails due to poor lighting conditions. The second scenario shows a person cooking on the stove. Video-captioning confuses the person's activity as washing dishes because, in the camera view, the person looks as if he were near a sink full of dishes. In contrast, \name\ generates a correct caption because it can extract 3D information from RF signals, and hence can tell that the person is near the stove not the sink.

To evaluate \name, we collect a captioning dataset of RF signals and floormaps, as well as the synchronized RGB videos.
Our experimental results demonstrate that: 1) \name~can obtain comparable results to video-captioning in visible scenarios. Specifically, on our test set, \name\ achieves 41.5 average BLEU and 26.7 CIDEr, while RGB-based video-captioning achieves 41.1 average BLEU and 27.0 CIDEr. 2) \name~continues to work effectively in dark and occluded conditions, where video-captioning methods fail. 3) Finally, our ablation study shows that the integration of the floormaps into the model and the multi-modal feature alignment both contribute significantly to improving performance. 

Finally, we summarize our contributions as follows:
\begin{Itemize}
    \item We are the first to caption people's daily-life at home, in the presence of bad lighting and occlusions. 
    \item We also introduce new modalities: the combination of RF and floormap, as well as new representations for both modalities that better tie them together.      
    \item We further introduce a multi-modal feature alignment training strategy for knowledge transfer from a video-captioning model to \name. 
    \item We evaluate our RF-based model and compare its performance to past work on video-captioning.  Our results provide new insights into the strengths and weaknesses of these two types of inputs. 
\end{Itemize}

%% file: source/related-2.tex
\section{Related Work}
\vskip -5 pt
\noindent\vskip 0.06in {\bf (a)~RGB-Based Video Captioning.}
Early works on video-captioning are direct extensions of image captioning. They pool features from individual frames across time, and apply image captioning models to video-captioning~\cite{venugopalan2015translating}. Such models cannot capture temporal dependencies in videos. Recent approaches, e.g., sequence-to-sequence video-to-text (S2VT), address this limitation by adopting recurrent neural networks (RNNs)~\cite{venugopalan2015sequence}. In particular, S2VT customizes LSTM for video-captioning and generates natural language descriptions using an encoder-decoder architecture. 
Follow-up papers improve this model by introducing an attention mechanism \cite{yao2015describing,fakoor2016memory,long2018video}, leveraging hierarchical architectures \cite{baraldi2017hierarchical,yu2016videornn,hu2019hierarchical,song2017hierarchical,gan2017stylenet,gan2017semantic}, 
or proposing new ways to improve feature extraction from video inputs, such as C3D features \cite{yao2015describing} or trajectories \cite{wu2018interpretable}. There have also been attempts to use reinforcement learning to generate descriptions from videos \cite{ranzato2015sequence,pasunuru2017reinforced,wang2018video}, in which they use the REINFORCE algorithm to optimize captioning scores.

\noindent\vskip 0.06in {\bf (b)~Human Behavior Analysis with Wireless Signals.} 
Recently, there has been a significant interest in analyzing the radio signals that bounce off people's bodies to understand human movements and behavior. Past papers have used radio signals to track a person's location \cite{adib20143d}, extract 3D skeletons of nearby people \cite{zhao2018rf}, or do action classification \cite{li2019making}. 
To the best of our knowledge, we are the first to generate natural language descriptions of continuous and complex in-home activities using RF signals. Moreover, we introduce a new combined modality based on RF+Floormap and a novel representation that highlights the interaction between these two modalities, as well as a multi-modal feature alignment training scheme to allow RF-based captioning to learn from existing video captioning datasets. 


%% file: source/rf.tex
\section{RF Signal Preliminary}

In this work, we use a radio commonly used in prior works on RF-based human sensing~\cite{zhao2018rf,hsu2017zero,hsu2019enabling,lien2016soli,peng2016fmcw,tian2018rf,chetty2017low,zhao2016emotion,zhao2019through,zhang2018latern,zhao2017learning,fan2020learning,hsu2017extracting}. The radio has two antenna arrays organized vertically and horizontally, each equipped with 12 antennas. The antennas transmit a waveform called FMCW \cite{stove1992linear} and sweep the frequencies from 5.4 to 7.2 GHz. Intuitively, the antenna arrays provide angular resolution and the FMCW provides depth resolution.

\label{sec:rf}
\begin{figure}[t]
\centering
\includegraphics[width=1.0\linewidth]{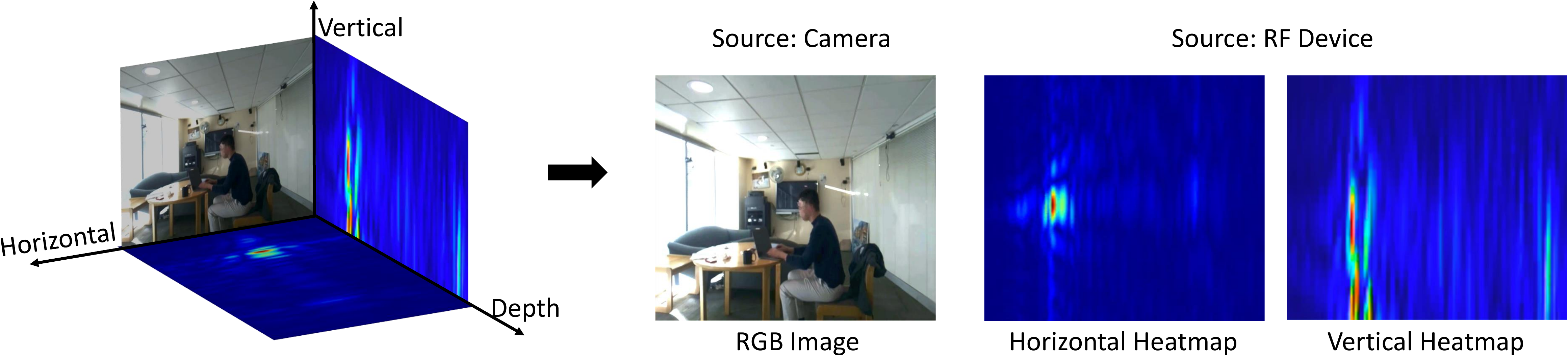}
\caption{\footnotesize{RF heatmaps and an RGB image recorded at the same time.}}\label{fig:rf-heatmaps}
\vskip -5 pt
\end{figure}

Our input RF signal takes the form of two-dimensional heatmaps, one from the horizontal array and the other from the vertical array. As shown in Figure \ref{fig:rf-heatmaps}, the horizontal heatmap is similar to a depth heatmap projected on a plane parallel to the ground, and the vertical heatmap is similar to a depth heatmap projected on a plane perpendicular to the ground. Red parts in the figure correspond to large RF power, while blue parts correspond to small RF power.  The radio generates 30 pairs of heatmaps per second. 

RF signals are  different from vision data. They contain much less information than RGB images. This is because the wave-length of RF signals is few centimeters making it hard to capture objects' shape using RF signals; they may even totally miss small objects such as a pen or cellphone. However, the FMCW radio enables us to get a relatively high resolution on depth ($\sim$8cm), making it much easier to locate a person. We harness this property to better associate RF signals and floormaps in the same coordinate system.

%% file: source/method-2.tex
\section{\name}
\label{sec:method}

\begin{figure*}[t]
\begin{center}
\includegraphics[width=1.0\linewidth]{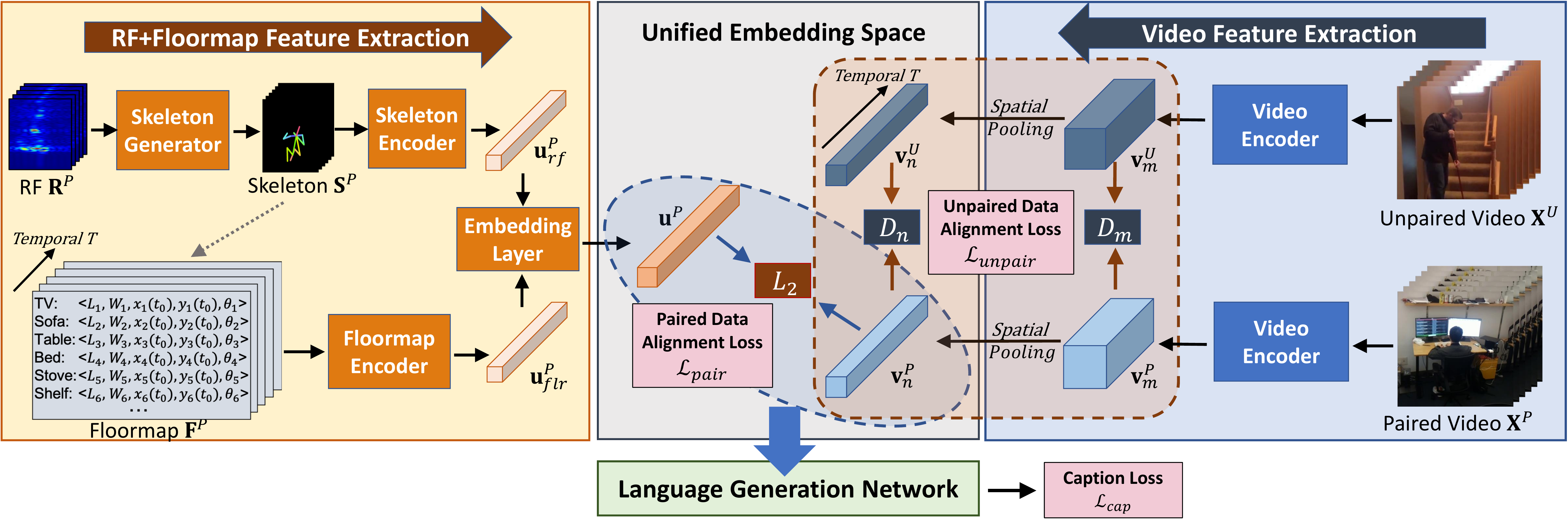}
\end{center}
\vskip -5 pt
\caption{\footnotesize{Model architecture. It contains four parts: RF+Floormap feature extraction (the left yellow box), video feature extraction (the right blue box), unified embedding space for feature alignment (the center grey box), and language generation network (the bottom green box). \name\ extracts features from RF signals and floormaps and combines them into a unified human-environment feature map. The features are then taken by the language generation network to generate captions. \name\ also extracts features from both paired videos (synchronized videos with RF+Floormap) and unpaired videos (an existing video captioning dataset), and gets the video representation. These features are used to distill knowledge from existing video dataset to \name. During training, \name\ uses the caption loss and the feature alignment loss to train the network. During testing, \name\ takes only the RF+Floormap without videos as input and generates captions.}}
\label{fig:method}
\vskip -15 pt
\end{figure*}

\vskip -5 pt
\name\ generates event captions using RF signals and floormaps. As shown in \autoref{fig:method}, our model first performs feature extraction from RF signals and floormaps, then combine them into a unified feature (the left yellow box). The combined feature is taken by a language generation network to generate captions (the bottom green box). Below, we describe the model in detail. We also provide implementation details in Appendix A.


\subsection{RF Signal Encoding}
RF signals have properties totally different from visible light, which are usually not interpretable by human. Therefore, it can be hard to directly generate captions from RF signals. However, recent works demonstrate that it is possible to generate accurate human skeletons from radio signals \cite{zhao2018rf}, and that the human skeleton is a succinct yet informative representation for human behavior analysis \cite{li2018co,du2015skeleton}. In this work, we first generate 3D human skeletons from RF signals, then extract the feature representations of the 3D skeletons.

Thus, the first stage in \name\ is a skeleton generator network, which has an 
architecture similar to the one in \cite{zhao2018rf}, with 90-frame RF heatmaps (3 seconds) as input; we refer to these 90 frames as an RF segment.  The skeleton generator first extracts information from the RF segment with a feature extraction network (12-layer ResNet). This is  followed by a region proposal network (6-layer ResNet) to generate region proposals for potential human bounding boxes and a pose estimation network (2-layer ResNet) to generate the final 3D skeleton estimations based on the feature maps and the proposals.  Note that these are dynamic skeletons similar to skeletons extracted from video segment. 

After we obtain the 3D skeletons from RF signals, we extract the feature representation through a skeleton encoder from each skeleton segment $ \textbf{S} $. The skeleton encoder is a Hierarchical Co-occurrence Network (HCN) \cite{li2018co}, which is a CNN-based network architecture for skeleton-based action recognition. We use the features from the last convolutional layer of HCN, denoted as $ \textbf{u}_{rf} $, as the encoded features for RF signals.

\begin{figure}[htbp]
\begin{center}
\includegraphics[width=0.81\linewidth]{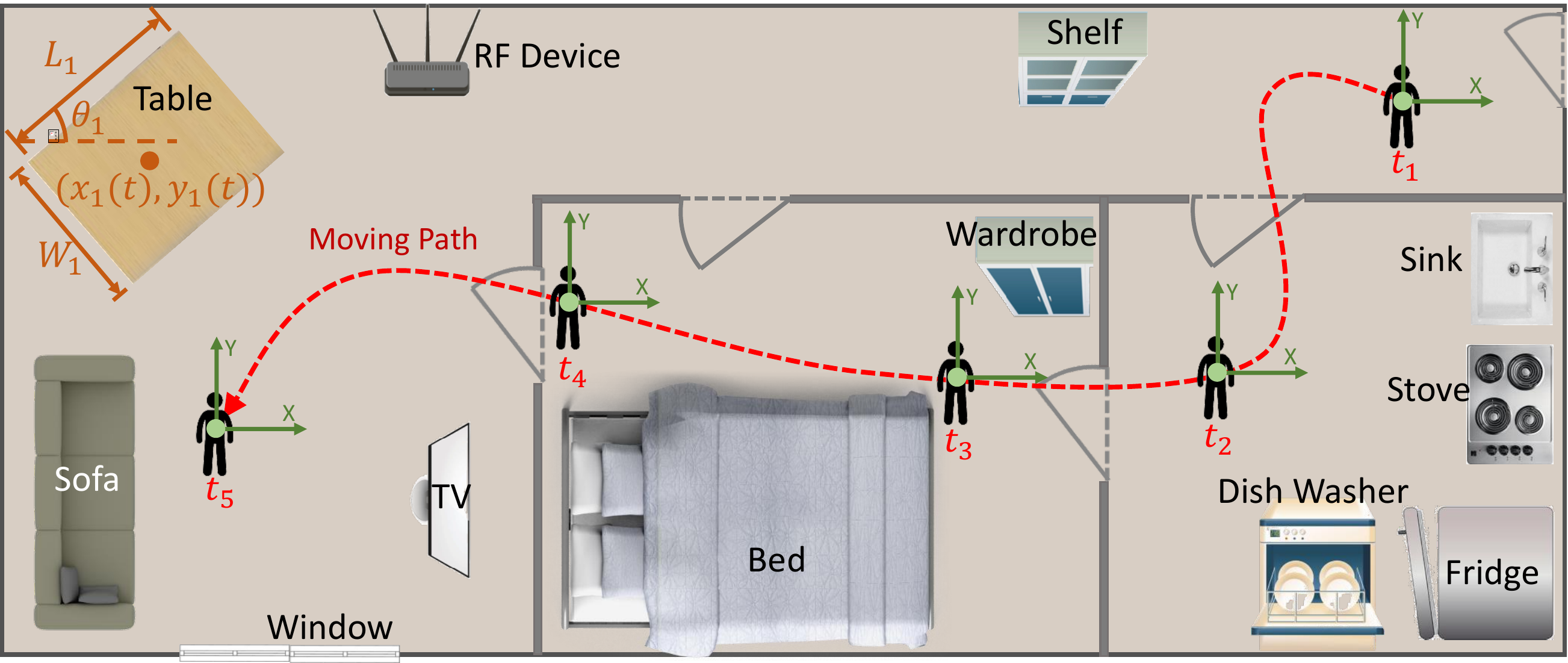}
\end{center}
\caption{\footnotesize{Illustration of floormap representation. Noted that this figure is not the input to our model but only a visualization. Red dotted line in the apartment denotes the moving path of a person from time $ t_{1} $ to $ t_{5}$. Green axes X-Y centered at the person illustrate our person-centric coordinate system, where the origin of the coordinate system is changed through time along with people's location. Under this person-centric coordinate system, at $ t^{th} $ time step, we describe each object using a 5-element tuple: (length $L$, width $W$, center coordinates $x(t), y(t)$, and rotation $\theta$), as exemplified using the \textbf{Table} in the figure.}}
\label{fig:floormap}
\vskip -15 pt
\end{figure}

\vskip -0.2 in\noindent
\subsection{Floormap Encoding}
\label{sec:floormap}

Many objects are transparent or semi-transparent to RF signals and act in a manner similar to air~\cite{adib2013see}. Even objects that are not transparent to RF signals, they can be partially invisible; this is because they can deflect the incident RF signal away from the emitting radio, preventing the radio from sensing them~\cite{zhao2018through,zhao2018rf}. 
Thus, to allow the model to understand interactions with objects, we must provide additional information about the surrounding environment. But we do not need to have every aspect of the home environment since most of the background information, e.g. the color or the texture of furniture, is irrelevant to captioning daily life. Therefore, we represent the in-home environment using the locations and sizes of objects -- the floormap. The floormap is easily measured with a laser meter. Once measured, the floormap tends to remain valid for a long time.  In our model, we only consider static objects relevant to people's in-home activities, e,g., bed, sofa, stove, or TV. To demonstrate the ease of collecting such measurements, we have asked 5 volunteers to measure the floormap for each of our test environments. The average time to measure one environment is about 7 mins.
 
Let $M$ be the number of objects, $N$ be the maximum number of instances of an object, then the floormap can be represented by a tensor $f\in\mathbb{R}^{M\times N\times O}$, where $O$ denotes the dimension for the location and size of each object, which is typically 5, i.e., length $L$, width $W$, the coordinate of the center $x(t),y(t)$, and its rotation $\theta$. 

Since people are more likely to interact with objects close to them,  we set the origin point of the floormap reference system to the location of the 3D skeleton extracted from RF. Specifically, we use a person-centric coordinate system as shown in the green $X$-$Y$ coordinates in \autoref{fig:floormap}. A person is moving around at home in a red-dotted path from time $ t_{1} $ to $ t_{5}$. At each time step $t_{i} $, we set the origin of the 3D coordinate system to be the center of the 3D human skeleton and the $X$-$Y$ plane to be parallel to the floor plane. The orientation of the $X$-axis is parallel to the RF device and $Y$-axis is perpendicular to the device. Each object is then projected onto the $X$-$Y$ plane. For example, at time $ t_{i} $, the center coordinates of the \textbf{Table} is $ (x_{1}(t_{i}), y_{1}(t_{i}))$, while its width, length and rotation $ (L_{1}, W_{1}, \theta_{1}) $ are independent of time. The floormap at time $ t_{i} $ is thus generated by describing each object $ k $ using a 5-element tuple: $(L_{k}, W_{k}, x_{k}({t_{i}}), y_{k}({t_{i}}),\theta_{k}) $, as shown in the left yellow box in Figure~\ref{fig:method}. In this way, each object's location is encoded w.r.t. the person's location, allowing our model to pay different attention to objects depending on their proximity to the person at that time instance. 

To extract features of the floormaps $ \textbf{F} $, we use a floormap encoder which is a two-layer fully-connected network which generates the encoded features for floormaps $ \textbf{u}_{flr} $. 

Using the encoded RF signal features $ \textbf{u}_{rf} $ and floormap features $ \textbf{u}_{flr} $, we generate a unified human-environment feature: $$ \textbf{u} = \psi(\textbf{u}_{rf} \oplus \textbf{u}_{flr}),$$ 
where $\oplus$ denotes the concatenation of vectors, and $\psi$ denotes an encoder to map the concatenated features to a joint embedding space. Here we use another two-layer fully-connected sub-network for $\psi$.

\subsection{Caption Generation}
To generate language descriptions based on the extracted features, we use an attention-based sequence-to-sequence LSTM model similar to the one in \cite{venugopalan2015sequence,wang2018video}. During the encoding stage, given the unified human-environment feature $\textbf{u}=\{u_t\}_{1}^{T} $ with time dimension $ T $, the encoder LSTM operates on its time dimension to generate hidden states $\{h_t\}_{1}^{T}$ and outputs $\{o_t\}_{1}^{T}$. During the decoding stage, the decoder LSTM uses $h_T$ as an initialization for hidden states and take inputs of previous ground-truth words with an attention module related to  $ \{u_t\}_{1}^{T} $, to output language sequence with $m$ words $\{w_1, w_2, ..., w_m\}$. The event captioning loss $\mathcal{L}_{cap}(\textbf{u})$ is then given by a negative-log-likelihood loss between the generated and the ground truth caption similar to \cite{venugopalan2015sequence,wang2018video}.

\section{Multi-Modal Feature Alignment Training}
Training \name\ requires a large labeled RF captioning dataset. Collecting such a dataset would be a daunting task. To make use of the existing large video-captioning dataset (e.g., Charades), we use a multi-model feature alignment strategy in the training process to transfer knowledge from video-captioning to  \name. However, RGB videos from Charades and RF signals have totally different distributions both in terms of semantics and modality. To mitigate the gap between them and make the knowledge distillation possible, we collect a small dataset where we record synchronized videos and RF from people performing activities at home. We also collect the floormaps and provide the corresponding natural language descriptions (for dataset details see section~\ref{sec:dataset}(a)). The videos in the small dataset are called paired videos, since they are in the same semantic space as their corresponding RF signal, while the videos in large existing datasets are unpaired videos with the RF signals. Both the paired and unpaired videos are in the same modality. Since the paired videos share the same semantics with the RF data, and the same modality with the unpaired videos, they can work as a bridge between video-captioning and \name, and distill knowledge from the video data to RF data. 


Our multi-modal feature alignment training scheme operates by aligning the features from RF+Floormaps and those from RGB videos. During training, our model extracts features from not only RF+Floormaps, but also paired and unpaired RGB videos, as shown in Figure~\ref{fig:method} (the right blue box). Besides the captioning losses, we add additional paired-alignment loss between paired videos and RF+Floormaps and unpaired-alignment loss between paired and unpaired videos. This ensures features from the two modalities are mapped
to a unified embedding space (the center grey box). Below, we describe the video encoder and the feature alignment method in detail.

\subsection{Video Encoding}
\label{sec:video_encoding}
We use the I3D model~\cite{carreira2017quo} pre-trained on Kinetics dataset to extract the video features. For each 64-frame video segment, we extract the Mixed\_5c features from I3D model, denoted as $ v_{m} $. We then apply a spatial average pooling on top of the Mixed\_5c feature and get the spatial pooled video-segment feature $ v_{n}$. For a video containing $ T $ non-overlapping video-segment, its Mixed\_5c features and the spatial-pooled features are denoted as $ \textbf{v}_{m} = \{v_{m}(t)\}_{1}^{T}$ and $ \textbf{v}_{n} = \{v_{n}(t)\}_{1}^{T}$. Therefore, the extracted features of paired videos $ \textbf{X}^{P} $ and unpaired videos $ \textbf{X}^{U} $ are denoted as $ \textbf{v}_{m}^{P} $, $ \textbf{v}_{n}^{P} $ and $ \textbf{v}_{m}^{U} $, $ \textbf{v}_{n}^{U} $. We use the spatial-pooled features to generate captions through the language generation model. The corresponding captioning loss is $\mathcal{L}_{cap}(\textbf{v}_{n}^{P})$ and $\mathcal{L}_{cap}(\textbf{v}_{n}^{U})$. 



\subsection{Alignment of Paired Data}
\label{sec:pair}
Since the synchronized video and RF+Floormap correspond to the exact same event, we use $L_2$ loss to align the features from paired video $ \textbf{v}_{n}^{P} $ in Sec~\ref{sec:video_encoding} and RF+Floormap $ \textbf{u}^{P} $ in Sec~\ref{sec:floormap} (we deonte a $ P $ here to indicate the paired data) to be consistent with each other in a unified embedding space, i.e., the paired data alignment loss $ \mathcal{L}_{pair}(\textbf{u}^{P}, \textbf{v}_{n}^{P}) = \left\lVert \textbf{u}^{P} - \textbf{v}_{n}^{P} \right\rVert_{2} $. 



\subsection{Alignment of Unpaired Data}
Existing large video-captioning datasets have neither synchronized RF signal nor the corresponding floormaps, so we cannot use the $L_2$-norm for alignment. Since we collect a small paired dataset, we can first train a video-captioning model on both paired and unpaired datasets, and then use the paired dataset to transfer knowledge to \name. However, the problem is that since the paired feature alignment is only applied on the paired dataset, the video-captioning model may overfit to the paired dataset and cause inconsistency between the distribution of features from paired and unpaired datasets.
To solve this problem, we align the paired and unpaired datasets by making the two feature distributions similar. We achieve this goal by applying discriminators on different layers of video features that enforces the video encoder to generate indistinguishable features given $ \textbf{X}^{P}$ and  $ \textbf{X}^{U} $. Specifically, we use two different layers of video features, i.e., $ \textbf{v}_{m} $ and $ \textbf{v}_{n} $ in Sec~\ref{sec:video_encoding}, to calculate the discriminator losses $\mathcal{L}_{unpair}(\textbf{v}_{m}^{P}, \textbf{v}_{m}^{U}) $ and $ \mathcal{L}_{unpair}(\textbf{v}_{n}^{P}, \textbf{v}_{n}^{U}) $. Since features from the paired videos are also aligned with the RF+Floormap features, this strategy effectively aligns the feature distribution of the unpaired video with the feature distribution of RF+Floormaps. The total loss of training process is shown as below:
\vskip -7 pt
\begin{eqnarray*}
\mathcal{L} &=& \mathcal{L}_{cap}(\textbf{u}^{P}) + \mathcal{L}_{cap}(\textbf{v}_{n}^{P}) + \mathcal{L}_{cap}(\textbf{v}_{n}^{U})  \\
&&  + \mathcal{L}_{pair}(\textbf{u}^{P}, \textbf{v}_{n}^{P})  \\
&&  + \mathcal{L}_{unpair}(\textbf{v}_{n}^{P}, \textbf{v}_{n}^{U}) + \mathcal{L}_{unpair}(\textbf{v}_{m}^{P}, \textbf{v}_{m}^{U}).
\end{eqnarray*}

%% file: source/experiment-3.tex
\section{Experiments}
\textbf{(a)~Datasets:~~}
\label{sec:dataset}
We collect a new dataset named RF Captioning Dataset (RCD). It provides synchronized RF signals, RGB videos, floormaps, and human-labeled captions to describe each event. We generate floormaps using a commercial laser meter. The floormaps are marked with  the following objects: cabinet, table, bed, wardrobe, shelf, drawer, stove, fridge, sink, sofa, television, door, window, air conditioner, bathtub, dishwasher, oven, bedside table. We use a radio device to collect RF signals, and a multi-camera system to collect multi-view videos, as the volunteers perform the activities. The synchronized radio signals and videos have a maximum synchronization error of 10 ms. The multi-camera system has 12 viewpoints to allow for accurate captioning even in cases where some viewpoints are occluded or the volunteers walk from one room to another room. 

To generate captions, we follow the method used to create the Charades dataset~\cite{sigurdsson2016hollywood} --i.e., we first  generate instructions similar to those used in Charades, ask the volunteers to perform activities according to the instructions, and record the synchronized RF signals and multi-view RGB videos. We then provide each set of multi-view RGB videos to Amazon Mechanical Turk (AMT) workers and ask them to label 2-4 sentences as the ground-truth language descriptions. 

\begin{table}[htbp]
\centering
\begin{tabular}{c|c|c|c|c|c|c|c|c}\hline
                           \#environments & \#clips & avg len & \#action types & \#object types & \#sentences & \#words & vocab & len (hrs) \\
\hline
10 &  1,035 & 31.3s & 157  & 38  & 3,610  & 77,762   & 6,910   & 8.99  \\
\hline    
\end{tabular}
\vskip 10 pt
\caption{Statistics of our RCD dataset.}
\label{tab:dataset}
\vskip -20 pt
\end{table}

We summarize our dataset statistics in \autoref{tab:dataset}. In total, we collect 1,035 clips in 10 different in-door environments, including bedroom, kitchen, living room, lounge, office, etc. Each clip on average spans 31.3 seconds. The RCD dataset exhibits two types of diversity. 
{\it 1. Diversity of actions and objects:} Our RCD dataset contains 157 different actions and 38 different objects to interact with. The actions and objects are the same as the Charades dataset to ensure a similar action diversity. The same action is performed at different locations by different people, and different actions are performed at the same location. For example, all of the following actions are performed in the bathroom next to the sink: brushing teeth, washing hands, dressing, brushing hair, opening/closing a cabinet, putting something on a shelf, taking something off a shelf, washing something,  etc.
{\it  2. Diversity of environments:} Environments in our dataset differ in their floormap, position of furniture, and the viewpoint of the RF device. Further, each environment and all actions performed in that environment are included either in testing or training, but not both.

\vskip 0.06in\noindent
\textbf{(b)~Train-test Protocol:~~}To evaluate \name~under visible scenarios, we do a 10-fold cross-validation on our RCD Dataset. Each time 9 environments are used for training, and the other 1 environment is used for testing. We report the average performance of 10 experiments.
To show the performance of \name~under invisible scenarios, e.g., with occlusions or poor lighting conditions, we randomly choose 3 environments (with 175 clips) and collect corresponding clips under invisible conditions. Specifically, in these 3 environments, we ask the volunteers to perform the same series of activities twice under the visible and invisible conditions (with the light on and off, or with an occlusion in front of the RF device and cameras), respectively. Later we provide the same ground truth language descriptions for the clips under invisible conditions as the corresponding ones under visible conditions. During testing, clips under invisible scenarios in these 3 environments are used for testing, and clips in the other 7 environments are used to train the model.



During training, we only use RGB videos from 3 cameras with good views instead of all 12 views in the multi-modal feature alignment between the video-captioning model and \name~model. Using multi-view videos will provide more training samples and help the feature space to be oblivious to the viewpoint. When testing the video-captioning model, we use the video from the master camera as it covers most of the scenes. The master camera is positioned atop of the RF device for a fair comparison.


We leverage the Charades caption dataset \cite{sigurdsson2016hollywood,wang2018video} as the unpaired dataset to train the video-captioning model. This dataset provides captions for different in-door human activities. It contains 6,963 training videos, 500 validation videos, and 1,760 test videos. Each video clip is annotated with 2-5 language descriptions by AMT workers.

\vskip 0.06in\noindent
\textbf{(c)~Evaluation Metrics:~~}
We adopt 4 caption evaluation scores widely used in video-captioning: BLEU \cite{papineni2002bleu}, METEOR \cite{denkowski2014meteor}, ROUGE-L \cite{lin2004rouge} and CIDEr \cite{vedantam2015cider}. BLEU-$n$ analyzes the co-occurrences of $n$ words between the predicted and ground truth sentences. METEOR compares exact token matches, stemmed tokens, paraphrase matches, as well as semantically similar matches using WordNet synonyms. ROUGE-L measures the longest common subsequence of two sentences. CIDEr measures consensus in captions by performing a Term Frequency Inverse Document Frequency (TF-IDF) weighting for each $n$ words. According to \cite{sigurdsson2016hollywood}, CIDEr offers the highest resolution and most similarity with human judgment on the captioning task for in-home events. We compute these scores using the standard evaluation code provided by the Microsoft COCO Evaluation Server \cite{chen2015microsoft}. All results are obtained as an average of 5 trials. We denote B@n, M, R, C short for BLEU-n, METEOR, ROUGE-L, CIDEr.

\subsection{Quantitative Results}
We compare \name~with state-of-the-art video-captioning baselines~\cite{hu2019hierarchical,venugopalan2015sequence,yao2015describing,fakoor2016memory}. The video-based models are trained on RGB data from both the Charades and RCD training sets, and tested on the RGB data of the RCD test set. \name~is trained on RF and floormap data from the RCD training sets. It also uses the RGB data from Charades and RCD training sets in multi-modal feature alignment training. It is then tested on the RF and floormap data of the RCD test set.

The results on the left side of \autoref{tab:result-main} show that \name~achieves comparable performance to state-of-the-art video captioning baselines in visible scenarios. The right side of \autoref{tab:result-main} shows that \name~also generates accurate language descriptions when the environment is dark or with occlusion, where video-captioning fails completely. The little reduction in \name's performance from the visible scenario is likely due to that occlusions attenuate RF signals and therefore introduce more noise in the RF heatmaps.

\begin{table}[htbp]
\centering
\vskip -10 pt
      \centering
      \begin{tabular}{l@{\hspace{0.2cm}}c@{\hspace{0.1cm}}c@{\hspace{0.1cm}}c@{\hspace{0.1cm}}c@{\hspace{0.1cm}}c@{\hspace{0.1cm}}c@{\hspace{0.1cm}}c@{\hspace{0.1cm}}c@{\hspace{0.1cm}}c@{\hspace{0.1cm}}c@{\hspace{0.1cm}}c@{\hspace{0.1cm}}c@{\hspace{0.1cm}}c@{\hspace{0.1cm}}c}
      \hline
       \multirow{2}{*}{Methods}  & \multicolumn{7}{c}{Visible Scenario} & \multicolumn{7}{c}{Dark/Occlusion Scenario}\\ \cmidrule(lr){2-8} \cmidrule(lr){9-15} & B@1 & B@2 & B@3 & B@4 & M & R & C & B@1 & B@2 & B@3 & B@4 & M & R & C \\
      \hline\hline
       S2VT \cite{venugopalan2015sequence} & 57.3 & 40.4 &  27.2 & 19.3 & 19.8 & 27.3 & 18.9 &  -  & -  & -  & -  & -   & -  & – \\
       SA \cite{yao2015describing} & 56.8 & 39.2 & 26.7 & 19.0 & 18.1 & 25.9 & 22.1 &  -  & -  & -  & -  & -   & -  & –\\ 
       MAAM \cite{fakoor2016memory} & 57.8 & 41.9 & 28.2 & 19.3 & 20.7 & 27.1 & 21.2 &  -  & -  & -  & -  & -   & -  & – \\
       HTM \cite{hu2019hierarchical} & 61.3  & 44.6  &  32.2  & 22.1  & 21.3   & 28.3   & 26.5 &  -  & -  & -  & -  & -   & -  & –  \\

       HRL \cite{wang2018video} &  \textbf{62.5}  & 45.3  &  32.9  & \textbf{23.8}  & \textbf{21.7}   & 28.5   & \textbf{27.0} &  -  & -  & -  & -  & -   & -  & –  \\
      \hline
      \textbf{\name}  & 62.3  & \textbf{45.9}  &\textbf{33.9}  & 23.5  & 21.1   & \textbf{28.9}   & 26.7  & \textbf{61.5}  & \textbf{45.5}  & \textbf{33.1}  & \textbf{22.6}  & \textbf{21.1}   & \textbf{28.3}   & \textbf{25.9} \\
      \hline
\end{tabular}


\caption{\footnotesize{Quantitative results for \name~and video-based captioning models. All models are trained on Charades and RCD training set, and tested on the RCD test set. The left side of the Table shows the results under visible scenarios, and the right side of the Table shows the results under scenarios with occlusions or without light.}}
\label{tab:result-main}
\vskip -20 pt
\end{table}

\vskip -0.2 in\noindent
\subsection{Ablation Study}
We conduct several ablation studies to demonstrate the necessity of each component in \name. All experiments here are evaluated on the visible test set of RCD.

\textbf{3D Skeleton vs. Locations:} One may wonder whether simply knowing the location of the person is enough to generate a good caption. This could happen if the RCD dataset has low diversity, i.e., each action is done in a specific location. This is however not the case in the RCD dataset, where each action is done in multiple locations, and each location may have different actions. To test this point empirically, we compare our model which extracts 3D skeletons from RF signals with a model that extracts only people locations from RF. We also compare with a model that extracts 2D skeletons with no locations (in this case the floormap's coordinate system is centered on the RF device).

\begin{table}[htbp]
\centering
\vskip -10 pt
\begin{tabular}{l |@{\hspace{0.2cm}} c @{\hspace{0.2cm}} c @{\hspace{0.2cm}}c @{\hspace{0.2cm}}c @{\hspace{0.2cm}}c @{\hspace{0.2cm}}c @{\hspace{0.2cm}}c}\hline
Method                           & B@1 & B@2 & B@3 & B@4 & M & R & C \\
\hline

Locations                 & 52.0  & 37.4  & 24.3  & 17.2  & 15.7   & 22.1   & 19.1  \\
2D Skeletons              & 56.5  & 39.8  & 26.9  & 18.8  & 18.0   & 24.1   & 22.3  \\
3D Skeletons            & \textbf{62.3}  & \textbf{45.9}  & \textbf{33.9}  & \textbf{23.5}  & \textbf{21.1}   & \textbf{28.9}   & \textbf{26.7}  \\
\hline    
\end{tabular}
\vskip 5 pt
\caption{\footnotesize{Comparison between using different human representations.}}
\label{tab:result-skl}
\vskip -20 pt
\end{table}

\autoref{tab:result-skl} shows that replacing \textit{3D skeletons} with  \textit{locations} or \textit{2D skeletons} yields poor performance.  This is because  \textit{locations} do not contain enough information of the actions performed by the person, and \textit{2D skeletons} do not contain information of the person's position with respect to the objects on the floormaps. These results show that: 1) our dataset is diverse and hence locations are not enough to generate correct captioning, and 2) our choice of representation, i.e., \textit{3D skeletons}, which combines information about both the people's locations and poses provides the right abstraction to learn meaningful features for proper captioning. 

\textbf{Person-Centric Floormap Representation:} 
In this work, we use a person-centric coordinate representation for the floormap and its objects, as described in \autoref{sec:floormap}. 
What if we simply use the image of the floormap with the objects, and mark the map with the person's location at each time instance?  We compare this \textit{image-based floormap} representation to our person-centric representation in \autoref{tab:result-floormap}. We use ResNet-18 pre-trained on ImageNet to extract features from the floormap image. The result shows that the image representation of floormap can achieve better performance than not having the floormap, but still worse than our person-centric representation. This is because it is much harder for the network to interpret and extract features from an image representation, since the information is far less explicit than our person-centric coordinate-based representation. 

\begin{table}[h]
\centering
\vskip -10 pt
\begin{tabular}{@{\hspace{0.2cm}}l@{\hspace{0.2cm}} |@{\hspace{0.2cm}} c @{\hspace{0.2cm}} c @{\hspace{0.2cm}}c @{\hspace{0.2cm}}c @{\hspace{0.2cm}}c @{\hspace{0.2cm}}c @{\hspace{0.2cm}}c@{\hspace{0.2cm}}}\hline
Method                           & B@1 & B@2 & B@3 & B@4 & M & R & C \\
\hline
w/o floormap & 56.3  & 40.8  & 27.7  & 18.5  & 18.1   & 24.0   & 22.1 \\
image-based floormap & 60.1  & 43.9  & 31.5  & 21.6  & 20.1  & 26.7   & 24.4  \\
person-centric floormap& \textbf{62.3}  & \textbf{45.9}  & \textbf{33.9}  & \textbf{23.5}  & \textbf{21.1}   & \textbf{28.9}   & \textbf{26.7}  \\
person-centric floormap+noise& 61.6  & 45.8  & 33.7  & 23.4  & 21.0   & 28.7   & 26.5  \\

\hline     
\end{tabular}
\vskip 5 pt
\caption{\footnotesize{Performance of \name~with or without using floormap, with different floormap representations, and with gaussian noise.}}
\label{tab:result-floormap}
\vskip -20 pt
\end{table}

\begin{figure*}[t]
\begin{center}
\includegraphics[width=1\linewidth]{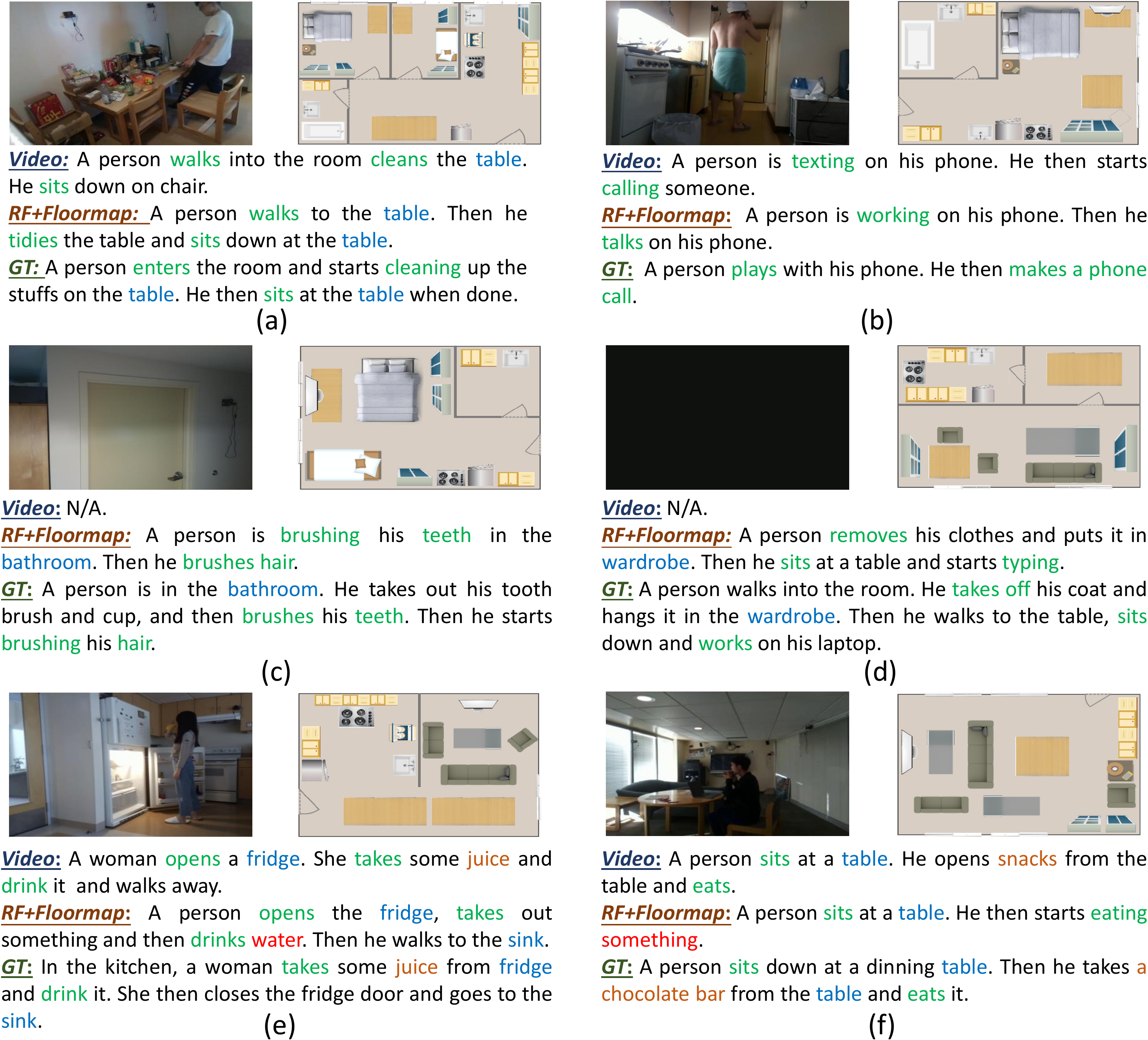}
\end{center}
\vskip -20 pt
\caption{\footnotesize{Examples from our RCD test set. Green words indicate actions. Blue words indicate objects included in floormap. Brown words indicate small objects not covered by floormap. Red words indicate the misprediction of small objects from \name. The first row shows \name~can generate accurate captions compared to the video-based captioning model under visible scenarios. The second row shows that \name~can still generate accurate captions when the video-based model does not work because of poor lighting conditions or occlusions. The third row shows the limitation of \name~that it may miss object color and detailed descriptions of small objects.}}\label{fig:results}
\vskip -15 pt
\end{figure*}

\textbf{Measurement Errors:} We analyze the influence of floormap measurement errors on our model's performance. We add a random gaussian noise with a 20cm standard deviation on location, 10cm on size and 30 degrees on object rotation. The results in the last row of \autoref{tab:result-floormap} show that the noise has very little effect on performance. This demonstrates that our model is robust to measurement errors.

\textbf{Feature Alignment}: Our feature alignment framework consists of two parts: the $L_2$-norm between paired dataset, and the discriminator between unpaired datasets. \autoref{tab:result-l2-dis}  quantifies the contribution of each of these alignment mechanisms to \name's performance. The results demonstrate that our multi-modal feature alignment training scheme helps \name~utilize the knowledge of the video-captioning model learned from the large video-captioning dataset to generate accurate descriptions, while training only on a rather small RCD dataset. We show a visualization of the features before and after alignment in the Appendix.

\begin{table}[h]
\centering
\vskip -10 pt
\begin{tabular}{l |@{\hspace{0.2cm}} c @{\hspace{0.2cm}} c @{\hspace{0.2cm}}c @{\hspace{0.2cm}}c @{\hspace{0.2cm}}c @{\hspace{0.2cm}}c @{\hspace{0.2cm}}c}\hline
Method                           & B@1 & B@2 & B@3 & B@4 & M & R & C \\
\hline
w/o $L_2$     & 52.5  & 38.0  & 25.7  & 18.5  & 16.6   & 23.1   & 20.3  \\
w/o discrim & 59.4  & 44.1  & 31.4  & 21.0  & 19.8  & 26.3  & 24.6 \\
\hline
\name  & \textbf{62.3}  & \textbf{45.9}  & \textbf{33.9}  & \textbf{23.5}  & \textbf{21.1}   & \textbf{28.9}   & \textbf{26.7}  \\
\hline     
\end{tabular}
\vskip 5 pt
\caption{\footnotesize{Performance of \name~network on RCD with or without $L_2$ loss and discriminator. Note that without adding the $L_2$ loss, \name~will not be affected by the video-captioning model. So if without the $L_2$ loss, then adding the discriminator loss on video-captioning model or not will not affect the \name's performance.}}
\label{tab:result-l2-dis}
\vskip -25 pt
\end{table}
%
%
%
%
%

\subsection{Qualitative Result}
In \autoref{fig:results}, we show six examples from the RCD test set. The first row under each image is the caption generated by state of the art video-based captioning model \cite{wang2018video}, the second row is the caption generated by \name, and the third row is the ground truth caption labeled by a human. 

The result shows that \name~can generate accurate descriptions of the person's activities (green) and interaction with the surrounding environment (blue), and continue to work well even in the presence of occlusions (\autoref{fig:results} (c)), and poor lighting (\autoref{fig:results} (d)). Video-based captioning is limited by bad lighting, occlusions and the camera's field of view. So if the person exits the field of view, video captioning can miss some of the events (\autoref{fig:results} (e)).

Besides poor lighting conditions, occlusions and field of view, video-captioning is also faced with privacy problems. For example, in \autoref{fig:results} (b), the person just took a bath and is not well-dressed. The video will record this content which is quite privacy-invasive. However, RF signal can protect privacy since it is not interpretable by a human, and it does not contain detailed information because of the relatively low resolution.


We also observe that \name\ has certain limitations. Since RF signals cannot capture details of objects such as color, texture, and shape, the model can mispredict those features. It can also mistake small objects. For example, in \autoref{fig:results} (e), the person is actually drinking orange juice, but \name\ predicts he is drinking water. Similarly, in \autoref{fig:results} (f), our model reports that the person is eating but cannot tell that he is eating a chocolate bar. The model also cannot distinguish the person's gender, so it always predicts ``he'' as shown in \autoref{fig:results} (e). 

\vskip -0.2 in\noindent
\vskip -5 pt
\subsection{Additional Notes on Privacy}
In comparison to images, RF signal is privacy-preserving because it is difficult to interpret by humans. However, one may also argue that since RF signals can track people though walls, they could create privacy concerns. This issue can be addressed through a challenge-response authentication protocol that prevent people from maliciously using RF signals to see areas that they are not authorized to access. More specifically, previous work~\cite{adib20143d} demonstrates that RF signals can sense human trajectories and locate them in space. Thus, whenever the user sets up the system to monitor an area, the system first challenges the user to execute certain moves (e.g., take two steps to the right, or move one meter forward), to ensure that the monitored person is the user. The system also asks the user to walk around the area to be monitored, and only monitors that area. Hence, the system would not monitor an area which the user does not have access to. 
\vskip -6 pt

%% file: source/conclusion.tex
\section{Conclusion}
In this paper, we introduce \name, a system that enables in-home daily-life captioning using RF signals and floormaps. 
We also introduce the combination of RF signal and floormap as new complementary input modalities, and propose a feature alignment training scheme to transfer the knowledge from large video-captioning dataset to \name. 
Extensive experimental results demonstrate that \name~can generate accurate descriptions of in-home events even when the environment is under poor lighting conditions or has occlusions.  We believe this work paves the way for many new applications in health monitoring and smart homes. 